\documentclass{article}

\usepackage{PRIMEarxiv}

\usepackage{graphicx}
\usepackage{amsmath}
\usepackage{xcolor}
\usepackage{todonotes}
\usepackage{multirow}
\usepackage{comment}
\usepackage{authblk}
\usepackage{todonotes}
\usepackage{float}

\usepackage[
  separate-uncertainty = true,
  multi-part-units = repeat
]{siunitx}

\usepackage{pgfplots}
\usepackage{cite}
\usepackage{tikz}
\usepackage{subfigure}
\usepackage{caption}
\usepackage{amsmath,amssymb,amsthm,float}

\title{ChatGPT or Human? Detect and Explain. Explaining Decisions of Machine Learning Model for Detecting Short ChatGPT-generated Text}
\author[1]{Sandra Mitrovi\'c}
\author[2]{Davide Andreoletti}
\author[2]{Omran Ayoub}
\affil[1]{Dalle Molle Institute for Artificial Intelligence - University of Southern Switzerland and University of Applied Sciences and Arts of Southern Switzerland, Switzerland}
\affil[2]{Information Systems and Networking Institute, University of Applied Sciences and Arts of Southern Switzerland, Switzerland}
\begin{document}

\date{}

\maketitle

\begin{abstract}
ChatGPT has the ability to generate grammatically flawless and seemingly-human replies to different types of questions from various domains. The number of its users and of its applications is growing at an unprecedented rate. Unfortunately, use and abuse come hand in hand. 
In this paper, we study whether a machine learning model can be effectively trained to accurately distinguish between original human and seemingly human (that is, ChatGPT-generated) text, especially when this text is short. Furthermore, we employ an explainable artificial intelligence framework to gain insight into the reasoning behind the model trained to differentiate between ChatGPT-generated and human-generated text. The goal is to analyze model's decisions and determine if any specific patterns or characteristics can be identified. Our study focuses on short online reviews, conducting two experiments comparing human-generated and ChatGPT-generated text. The first experiment involves ChatGPT text generated from custom queries, while the second experiment involves text generated by rephrasing original human-generated reviews. We fine-tune a Transformer-based model and use it to make predictions, which are then explained using SHAP. We compare our model with a perplexity score-based approach and find that disambiguation between human and ChatGPT-generated reviews is more challenging for the ML model when using rephrased text. However, our proposed approach still achieves an accuracy of 79\%. Using explainability, we observe that ChatGPT's writing is polite, without specific details, using fancy and atypical vocabulary, impersonal, and typically it does not express feelings. 

\end{abstract}

\section{Introduction}\label{Intro}

On November 30, OpenAI released ChatGPT \cite{ChatGPT}, a large language model that has shown unprecedented performance in understanding user queries and generating human-like text. Designed in a conversational format, ChatGPT caught such a tremendous attention that just within a few days upon its launch was put to the test by millions of users worldwide. The reasons for this are two-fold.

First, ChatGPT is powered by the latest progress in the natural language processing (NLP) field, building upon the advancements of its predecessors, GPT-3 series of language models \cite{brown2020language} and InstructGPT \cite{ouyang2022training}, and fixing their known problems. Specifically, the GPT-3 models inability to align with user needs in a helpful and safely manner, was overcome by introducing reinforcement learning from human feedback \cite{christiano2017deep} for training InstructGPT. Its reward model is trained according to the order of user preferences and the underlying GPT-3 model policy is fine-tuned to maximize this reward. 
This way, the model understands the meaning and the intent behind a user's query, which, consequently, makes it capable of responding in a most relevant and helpful way. To make the model safety-proof and avoid generation of offensive, inappropriate and simply wrong (fact-based) texts, which was InstructGPT major flaw, the training strategy for ChatGPT was empowered by augmenting the dataset with additional conversations where humans were acting both as humans and chatbots. This massive dataset allowed ChatGPT to learn the patterns and nuances of human language. As a result, it is able to generate a highly realistic text, almost indistinguishable from the one written by a human.

Second, one of the most impressive aspects of ChatGPT is its ample functionality range, that is, the ability to handle broad variety of query types that, additionally, relate to diverse domains. As such, ChatGPT can handle anything from simple information requests to more complex and open-ended questions. ChatGPT is seemingly able to successfully generate a human-like text just about anything. This makes it an ideal chatbot for a wide range of applications, such as customer service, e-commerce, and even creative writing. 

The rise of revolutionary AI-based chatbots, such as ChatGPT,
however, highlights the importance of the ability to detect if a text is generated by an AI or by a human. This can have serious implications for a variety of fields, including information security and digital forensics. For instance, in information security, the ability to detect AI-generated text is essential for identifying and protecting against malicious use of AI, such as the spread of misinformation and disinformation, or social engineering attacks. To ensure the trustworthiness and accuracy of information, it is important to develop methods for detecting AI-generated text, especially since it could be used in sensitive fields like political campaigns, financial reports, legal documents or customer reviews (such as customer reviews of products, restaurants or movies). 

In this work, we focus on the detection of ChatGPT-generated text for the case of restaurant reviews. We particularly focus on reviews as these typically do not pass through (rigorous) verification (unlike e.g., legal texts) even though their abuse could also have substantial financial impact. In particular, if used for generating false reviews in the restaurant business, it could either damage the reputation of a high-quality business or, on the other hand, falsely improve the reputation of low-quality business, potentially harming its clients. Additionally, we consider reviews that are relatively short and do not exceed two sentences, which is expected to make the detection of AI-generated text more challenging than in case of a longer text. As data, we use publicly-available human-generated reviews and we manually collect ChatGPT-generated reviews. The aim of our study is twofold: i) to evaluate the capability of a machine learning (ML)-based intelligent system to distinguish between text written by humans and text originated by ChatGPT and ii) to understand which characteristics of the text drive the intelligent system to detect AI-generated text. To this end, we first develop a Transformer-based ML approach to classify text as either generated by a human or by ChatGPT. We compare the performance of our proposed approach to a perplexity-based classification approach\footnote{Our motivation behind considering perplexity-based classification approach is based on recent developed applications that propose the use of perplexity score\cite{rosenfeld1996maximum} to detect potentially AI-generated content \cite{tian_2023}.}. Then, we apply Shapley Additive Explanations (SHAP), an eXplainable Artificial Intelligence (XAI) framework, to extract explanations of model's decisions aiming at uncovering model's reasoning. Experimental results show that both approaches show an acceptable to good performance in detecting ChatGPT-generated text and that the ML-based approach outperforms the perplexity-based classification approach. Furthermore, obtained explanations show that, despite this being a preliminary study, we can characterize ChatGPT review language as very formal, polite and impersonal, describing rather experiences instead of feelings, and focusing on general concepts without going into details. 

The paper is organized as follows: in Section \ref{sec:ProbMeth} we define the problem and corresponding research questions and explain in details our methodology; in Section \ref{sec:ExpSetup} we describe our datasets and experimental setup including benchmark; in Section \ref{sec:Results} we present obtained quantitative results and local explanations; in Section \ref{sec:Discussion} we discuss them and in \ref{sec:Conc} we conclude and provide some ideas for the future work.

\section{Detecting ChatGPT-generated Text}\label{sec:ProbMeth}

This section first presents the problem statement and the posed research questions and then describes the proposed methodology. 

\subsection{Problem Statement and Research Questions}
We frame the problem at hand as a supervised classification task, where the objective is to learn a mapping between a representation of the text and a binary variable, which is $1$ if the text is generated by ChatGPT, and $0$ otherwise. More formally, by means of ML strategies, we learn a function $f$ that, given an input text $t_{i}$, represented as a set of features $[f_{1}^{i},...,f_{k}^{i}]$, outputs an estimated label ${\hat{l}}_{i} \in \{0,1\}$, i.e., ${\hat{l}}_{i} \!\!= \!\!f\left(t_{i} \right)$. 
Furthermore, we are interested in explaining ML model's decisions and extracting insights that allow us to understand which features (e.g., words) contribute positively to detection of ChatGPT-generated text. 

To systematically discuss our findings, we pose the following research questions (RQs): \\
RQ1) To what extent can ML-based approach and perplexity-based approach detect AI-generated text of relatively short length? And if AI-generated text is a rephrase of the original, human-generated, text?\\
RQ2) What are the characteristics of text that contribute to detecting an AI-generated text? Can we extract insights by examining explanations of ML model decisions?

\begin{figure*}[t!]
    \centering
    \includegraphics[width=0.85\textwidth]{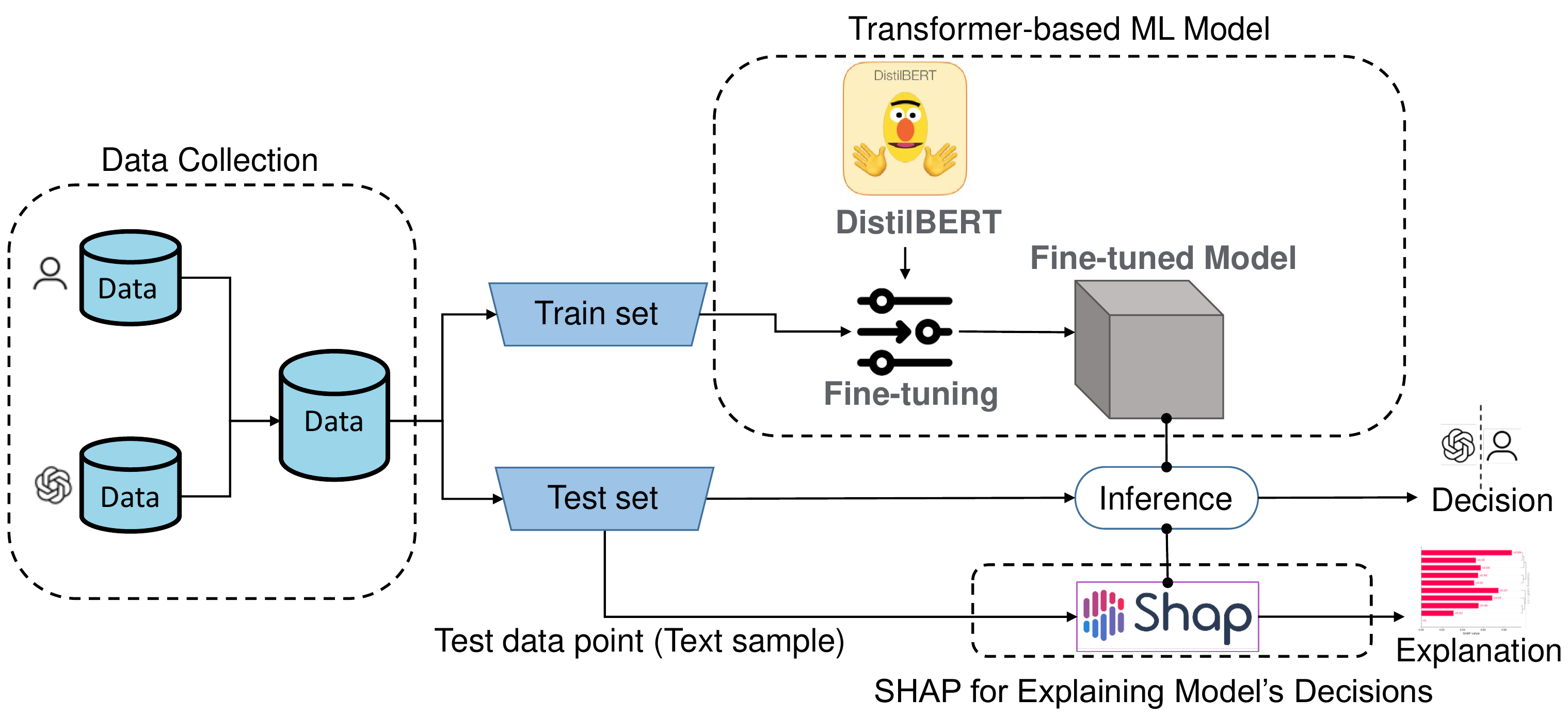}
    \caption{Schematic representation of the study design and building blocks.}
    \label{fig:methodology}
\end{figure*}

\subsection{Methodology}\label{Meth}

Our approach consists of two main building blocks (see Figure \ref{fig:methodology}). The first consists of a machine learning model trained to discriminate between text samples generated by a human and text samples generated by ChatGPT. The second is an explainable AI framework to explain and interpret outcomes of the ML model used for discrimination. We leverage XAI to delve into understanding the inner workings of the obtained model aiming to derive the insights about writing styles of humans, at the one hand, and that of ChatGPT, at the other, and their subsequent differences. In the following, we discuss these two building blocks in more details.

\textbf{Transformer-based ML model} Different directions could be pursued in order to extract useful features from a piece of text and perform text classification. Given the recent success of the Transformer-based architectures \cite{vaswani2017attention, devlin2018bert} for solving different NLP tasks, we decided to follow the same vein to build our classifier. As such, we start with a Transformer-based model pre-trained for the sequence classification task, using its corresponding tokenizer to pre-process data. We then fine-tune the model on the training subset of collected data to detect whether the text sample belongs to the positive (ChatGPT-generated) or the  negative (human-generated) class. Consequently, the fine-tuned model is used for inference on the testing subset. Finally, the obtained classification scores are evaluated against the ground truth.

For implementing the pre-trained model, we have used the uncased version of DistilBERT, a lightweighted model trained using BERT \cite{devlin2018bert} as a teacher. For fine-tuning of the pre-trained model, HuggingFace Transformers PyTorch Trainer class was used, leaving, however, the default hyper-parameters. The inference step (based on the fine-tuned model) was implemented using text classification pipeline of HuggingFace Transformers.

\textbf{SHAP for Explaining Model's Decisions}
We now discuss the XAI building block. As a XAI framework, we use Shapley Additive Explanations (SHAP) \cite{SHAP-NIPS2017}. 

SHAP is a model agnostic method for explaining the output of any ML model, and hence, helps in understanding the decision-making process of complex models, such as those used in text classification. In text classification, SHAP assigns a feature importance value to each input feature of a text classification model, and it does so in a way that provides a comprehensive, fair and unified measure of feature importance. This is achieved by using the Shapley values from cooperative game theory which supply a way to distribute a value among a group of players such that it is distributed fairly among the players depending on their contribution to the overall value. When applied to text classification models, SHAP can help identify which words or phrases in a given text are the most important in determining the model's output. This can help identify the specific parts of the text (words, or set of words) that are driving a particular classification outcome, which can in turn help improve the model's accuracy and interpretability.

In our analysis, we rely on local explanations, which refer to the process of providing an explanation for a specific decision and which are provided in the form of feature importance scores (SHAP value), indicating the relative importance of different input features in the model's decision. For the problem at hand, extracted local explanations show which words in a given text were most important in determining the model's output.

\section{Datasets and Experimental Setup}\label{Setup}
\label{sec:ExpSetup}

This section describes the datasets used and the experimental setup, in addition to the benchmark approach considered. 

\begin{table*}[t]
\small
\centering
\caption{Examples of queries submitted to ChatGPT, and their relative responses, for both ChatGPT-generated datasets}
\label{tab:mytable}
\begin{tabular}{|p{7.5cm}|p{7.5cm}|}
\hline
\multicolumn{2}{|c|}{\textbf{Samples from $ChatGPT_{query}$ dataset (response to custom queries)}} \\ 
\hline
\textbf{Prompt} & \textbf{ChatGPT Response} \\ 
\hline
\textit{write me a two-line review about a restaurant that has some good aspects} & \textit{I had a great experience at this restaurant! The food was delicious and the service was friendly and efficient.} \\ 
\hline
\textit{please write me a 3-lines review for a bad restaurant} & \textit{The food at this restaurant was a major letdown. Overcooked, bland, and unappealing. The service was slow and inattentive. I would not recommend this place to anyone.} \\ 
\hline
\textit{You are a client at a restaurant. Write a review about the restaurant you visited. Do not name of the restaurant. Do not start with "the". Alternate the focus of the review (e.g., food, drink, cost, politeness, place, music)} & \textit{Overall, my experience at this restaurant was unforgettable and worth the splurge. It was a truly special dining experience that exceeded all of my expectations.} \\ 
\hline
\textit{write me a review about a restaurant that has some good and some bad aspects} & \textit{I recently visited this restaurant and had a mixed experience. The food was generally delicious and beautifully presented, with standout dishes including the roast chicken and the sea bass. However, the service was incredibly slow and our server seemed uninterested and inattentive. Additionally, the atmosphere was quite noisy and chaotic, which detracted from the overall dining experience.} \\ 
\hline

\multicolumn{2}{|c|}{\textbf{Samples from $ChatGPT_{rephrase}$ dataset (rephrasing human-generated)}} \\ 
\hline
\textbf{Prompt (Human-generated reviews)} & \textbf{ChatGPT Response} \\ 
\hline
\textit{The selection on the menu was great and so were the prices.} & \textit{The menu had a great selection and the prices were good.} \\ 
\hline
\textit{Point your finger at any item on the menu, order it and you won't be disappointed.} & \textit{No matter what you order from the menu, you won't be disappointed.} \\ 
\hline
\textit{The one down note is the ventilation could use some upgrading.} & \textit{The only drawback at this restaurant is that the ventilation could be improved.} \\ 
\hline
\textit{I believe that this place is a great stop for those with a huge belly and hankering for sushi.} & \textit{I believe this place is a great stop for those with a big appetite and a desire for sushi.} \\ 
\hline
\end{tabular}
\label{tab:gpt_generated}
\end{table*}

\subsection{Datasets}
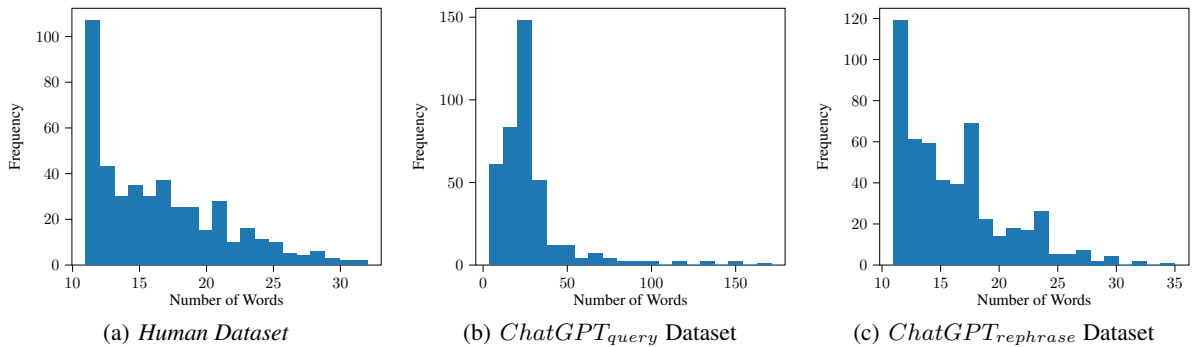
\begin{figure*}[b!]{
\centering
\subfigure[{\textit{Human Dataset}}]
{
\begin{tikzpicture}[scale=0.6]

\definecolor{darkgray176}{RGB}{176,176,176}
\definecolor{steelblue31119180}{RGB}{31,119,180}

\begin{axis}[
tick align=outside,
tick pos=left,
x grid style={darkgray176},
xlabel={Number of Words},
xmin=9.95, xmax=33.05,
xtick style={color=black},
y grid style={darkgray176},
ylabel={Frequency},
ymin=0, ymax=112.35,
ytick style={color=black}
]
\draw[draw=none,fill=steelblue31119180] (axis cs:11,0) rectangle (axis cs:12.05,107);
\addlegendimage{ybar,ybar legend,draw=none,fill=steelblue31119180}

\draw[draw=none,fill=steelblue31119180] (axis cs:12.05,0) rectangle (axis cs:13.1,43);
\draw[draw=none,fill=steelblue31119180] (axis cs:13.1,0) rectangle (axis cs:14.15,30);
\draw[draw=none,fill=steelblue31119180] (axis cs:14.15,0) rectangle (axis cs:15.2,35);
\draw[draw=none,fill=steelblue31119180] (axis cs:15.2,0) rectangle (axis cs:16.25,30);
\draw[draw=none,fill=steelblue31119180] (axis cs:16.25,0) rectangle (axis cs:17.3,37);
\draw[draw=none,fill=steelblue31119180] (axis cs:17.3,0) rectangle (axis cs:18.35,25);
\draw[draw=none,fill=steelblue31119180] (axis cs:18.35,0) rectangle (axis cs:19.4,25);
\draw[draw=none,fill=steelblue31119180] (axis cs:19.4,0) rectangle (axis cs:20.45,15);
\draw[draw=none,fill=steelblue31119180] (axis cs:20.45,0) rectangle (axis cs:21.5,28);
\draw[draw=none,fill=steelblue31119180] (axis cs:21.5,0) rectangle (axis cs:22.55,10);
\draw[draw=none,fill=steelblue31119180] (axis cs:22.55,0) rectangle (axis cs:23.6,16);
\draw[draw=none,fill=steelblue31119180] (axis cs:23.6,0) rectangle (axis cs:24.65,11);
\draw[draw=none,fill=steelblue31119180] (axis cs:24.65,0) rectangle (axis cs:25.7,10);
\draw[draw=none,fill=steelblue31119180] (axis cs:25.7,0) rectangle (axis cs:26.75,5);
\draw[draw=none,fill=steelblue31119180] (axis cs:26.75,0) rectangle (axis cs:27.8,4);
\draw[draw=none,fill=steelblue31119180] (axis cs:27.8,0) rectangle (axis cs:28.85,6);
\draw[draw=none,fill=steelblue31119180] (axis cs:28.85,0) rectangle (axis cs:29.9,3);
\draw[draw=none,fill=steelblue31119180] (axis cs:29.9,0) rectangle (axis cs:30.95,2);
\draw[draw=none,fill=steelblue31119180] (axis cs:30.95,0) rectangle (axis cs:32,2);
\end{axis}

\end{tikzpicture}
}
\subfigure[{$ChatGPT_{query}$ Dataset}]
{
\begin{tikzpicture}[scale=0.6]

\definecolor{darkgray176}{RGB}{176,176,176}
\definecolor{steelblue31119180}{RGB}{31,119,180}

\begin{axis}[
tick align=outside,
tick pos=left,
x grid style={darkgray176},
xlabel={Number of Words},
xmin=-4.35, xmax=179.35,
xtick style={color=black},
y grid style={darkgray176},
ylabel={Frequency},
ymin=0, ymax=155.4,
ytick style={color=black}
]
\draw[draw=none,fill=steelblue31119180] (axis cs:4,0) rectangle (axis cs:12.35,61);
\addlegendimage{ybar,ybar legend,draw=none,fill=steelblue31119180}

\draw[draw=none,fill=steelblue31119180] (axis cs:12.35,0) rectangle (axis cs:20.7,83);
\draw[draw=none,fill=steelblue31119180] (axis cs:20.7,0) rectangle (axis cs:29.05,148);
\draw[draw=none,fill=steelblue31119180] (axis cs:29.05,0) rectangle (axis cs:37.4,51);
\draw[draw=none,fill=steelblue31119180] (axis cs:37.4,0) rectangle (axis cs:45.75,12);
\draw[draw=none,fill=steelblue31119180] (axis cs:45.75,0) rectangle (axis cs:54.1,12);
\draw[draw=none,fill=steelblue31119180] (axis cs:54.1,0) rectangle (axis cs:62.45,4);
\draw[draw=none,fill=steelblue31119180] (axis cs:62.45,0) rectangle (axis cs:70.8,7);
\draw[draw=none,fill=steelblue31119180] (axis cs:70.8,0) rectangle (axis cs:79.15,4);
\draw[draw=none,fill=steelblue31119180] (axis cs:79.15,0) rectangle (axis cs:87.5,2);
\draw[draw=none,fill=steelblue31119180] (axis cs:87.5,0) rectangle (axis cs:95.85,2);
\draw[draw=none,fill=steelblue31119180] (axis cs:95.85,0) rectangle (axis cs:104.2,2);
\draw[draw=none,fill=steelblue31119180] (axis cs:104.2,0) rectangle (axis cs:112.55,0);
\draw[draw=none,fill=steelblue31119180] (axis cs:112.55,0) rectangle (axis cs:120.9,2);
\draw[draw=none,fill=steelblue31119180] (axis cs:120.9,0) rectangle (axis cs:129.25,0);
\draw[draw=none,fill=steelblue31119180] (axis cs:129.25,0) rectangle (axis cs:137.6,2);
\draw[draw=none,fill=steelblue31119180] (axis cs:137.6,0) rectangle (axis cs:145.95,0);
\draw[draw=none,fill=steelblue31119180] (axis cs:145.95,0) rectangle (axis cs:154.3,2);
\draw[draw=none,fill=steelblue31119180] (axis cs:154.3,0) rectangle (axis cs:162.65,0);
\draw[draw=none,fill=steelblue31119180] (axis cs:162.65,0) rectangle (axis cs:171,1);
\end{axis}

\end{tikzpicture}
}
\subfigure[{$ChatGPT_{rephrase}$ Dataset}]
{
\begin{tikzpicture}[scale=0.6]

\definecolor{darkgray176}{RGB}{176,176,176}
\definecolor{steelblue31119180}{RGB}{31,119,180}

\begin{axis}[
tick align=outside,
tick pos=left,
x grid style={darkgray176},
xlabel={Number of Words},
xmin=9.8, xmax=36.2,
xtick style={color=black},
y grid style={darkgray176},
ylabel={Frequency},
ymin=0, ymax=124.95,
ytick style={color=black}
]
\draw[draw=none,fill=steelblue31119180] (axis cs:11,0) rectangle (axis cs:12.2,119);
\addlegendimage{ybar,ybar legend,draw=none,fill=steelblue31119180}

\draw[draw=none,fill=steelblue31119180] (axis cs:12.2,0) rectangle (axis cs:13.4,61);
\draw[draw=none,fill=steelblue31119180] (axis cs:13.4,0) rectangle (axis cs:14.6,59);
\draw[draw=none,fill=steelblue31119180] (axis cs:14.6,0) rectangle (axis cs:15.8,41);
\draw[draw=none,fill=steelblue31119180] (axis cs:15.8,0) rectangle (axis cs:17,39);
\draw[draw=none,fill=steelblue31119180] (axis cs:17,0) rectangle (axis cs:18.2,69);
\draw[draw=none,fill=steelblue31119180] (axis cs:18.2,0) rectangle (axis cs:19.4,22);
\draw[draw=none,fill=steelblue31119180] (axis cs:19.4,0) rectangle (axis cs:20.6,14);
\draw[draw=none,fill=steelblue31119180] (axis cs:20.6,0) rectangle (axis cs:21.8,18);
\draw[draw=none,fill=steelblue31119180] (axis cs:21.8,0) rectangle (axis cs:23,17);
\draw[draw=none,fill=steelblue31119180] (axis cs:23,0) rectangle (axis cs:24.2,26);
\draw[draw=none,fill=steelblue31119180] (axis cs:24.2,0) rectangle (axis cs:25.4,5);
\draw[draw=none,fill=steelblue31119180] (axis cs:25.4,0) rectangle (axis cs:26.6,5);
\draw[draw=none,fill=steelblue31119180] (axis cs:26.6,0) rectangle (axis cs:27.8,7);
\draw[draw=none,fill=steelblue31119180] (axis cs:27.8,0) rectangle (axis cs:29,2);
\draw[draw=none,fill=steelblue31119180] (axis cs:29,0) rectangle (axis cs:30.2,4);
\draw[draw=none,fill=steelblue31119180] (axis cs:30.2,0) rectangle (axis cs:31.4,0);
\draw[draw=none,fill=steelblue31119180] (axis cs:31.4,0) rectangle (axis cs:32.6,2);
\draw[draw=none,fill=steelblue31119180] (axis cs:32.6,0) rectangle (axis cs:33.8,0);
\draw[draw=none,fill=steelblue31119180] (axis cs:33.8,0) rectangle (axis cs:35,1);
\end{axis}

\end{tikzpicture}
}
\caption{Distribution of the length of the text samples (in words) of each of the three data sets considered in our work: (a) Human, (b) $ChatGPT_{query}$ and $ChatGPT_{rephrase}$.}
\label{fig:dataset_stats}
}
\end{figure*}

We consider three datasets in our work, namely, a publicly available human-generated dataset containing actual restaurant reviews\footnote{https://www.kaggle.com/competitions/restaurant-reviews/overview}, and two datasets consisting of restaurant reviews generated by ChatGPT. The two ChatGPT-generated datasets will be made publicly available. Figure \ref{fig:dataset_stats} shows the distribution of the number of words in the reviews of each dataset. The three datasets are described in the following. 

\textbf{Human Dataset}: The human-generated dataset has been crafted for a Kaggle competition on sentiment analysis, and consists of $1000$ review texts and their corresponding sentiment labels. We discard those labels as they are not required for the purpose of our work. 
We performed minimal preprocessing of the dataset to filter the reviews with less than 11 words.

$\mathbf{ChatGPT_{query}}$ \textbf{Dataset}: The first ChatGPT-generated dataset, denoted as $ChatGPT_{query}$, has been collected by submitting custom queries to ChatGPT. This dataset consists of $395$ reviews, generated by the queries of various types, as the aim was to diversify the outputs of ChatGPT. For example, we drove it to focus on specific aspects, e.g. the cost and the quality of the food, or to produce reviews of varied lengths. Examples of queries and relative outputs of ChatGPT can be found in Table \ref{tab:gpt_generated}.

$\mathbf{ChatGPT_{rephrase}}$ \textbf{Dataset}: The second ChatGPT-generated dataset, denoted as $ChatGPT_{rephrase}$, has been collected by asking ChatGPT to rephrase each of the reviews of the human-generated dataset. Hence, this dataset consists of $1000$ reviews, each of which is a rephrased version of the original human reviews.
Examples of reviews we asked ChatGPT to rephrase, along with their relative outputs can be found in Table \ref{tab:gpt_generated}. \\

The distribution of the number of reviews per review length (measured in terms of number of words) for all three datasets can be seen in \ref{fig:dataset_stats}. Except for a minor number of cases for $ChatGPT_{query}$, the reviews are mostly quite short texts. 

\subsection{Experiments}
In this section, we explain our two experiments, designed to test discriminating power of our machine learning model. \\
\textbf{Experiment 1} Within the first experiment we use human-generated dataset and the first ChatGPT-generated dataset, $ChatGPT_{query}$, asking ChatGPT to generate reviews. \\
\textbf{Experiment 2}
In the second experiment, we again use human-generated dataset but we use instead $ChatGPT_{rephrase}$ dataset, consisting of rephrased human reviews. As we noticed that some rephrased reviews are actually pure copies of the original (human) reviews, these were initially filtered. Additionally, it is worth noticing that unless further filtering is performed, it might occur that the original version of a review is used for training while its rephrased version is used for testing (or vice versa), which might influence the classification performance. However, we intentionally decided not to intervene in order to verify whether, as we would expect, the rephrased setup with $ChatGPT_{rephrase}$ will still underperform the one with human reviews-unrelated queries $ChatGPT_{query}$.


\subsection{Benchmark Approach: Perplexity-based Classification}
In addition to our proposed ML-based classification approach, we develop a classification approach based on the \emph{perplexity score} \cite{rosenfeld1996maximum}. Typically, perplexity is used to evaluate the performance of language models in NLP tasks such as text generation and machine translation \cite{lample2018phrase,vaswani2018tensor2tensor,wang2019learning,lewis2019bart,geluykens2021neural}. It is a measure of how well a language model is able to predict a given text. Specifically, perplexity measures the uncertainty of generated text from probabilistic language models. A lower perplexity indicates that the model is better able to predict the text. In other words, texts with lower perplexities are more likely to be generated by language models while texts with higher perplexities are more likely to be generated by a human. 

We develop the perplexity-based classification approach following four steps. First, we split the dataset containing human- and ChatGPT-generated text samples into two sets (referred to as \emph{training} and \emph{testing} sets). Second, we evaluate the perplexity score of each sample in the training set using GPT-2. The idea is to find an optimal \emph{threshold perplexity value} (\emph{th}), upon which we would perform classification, assigning a label 0 (human) if the perplexity score of a text sample exceeds it (and 1 i.e. ChatGPT, otherwise). This value  \emph{th} is determined as to maximize the F1-score on the training set. Finally, we evaluate the perplexity score of the text samples in the test set and classify based on \emph{th} found in the previous step. 

\section{Quantitative Results and Explanations}
\label{sec:Results}

In this section we address the research questions posed earlier by discussing our experimental results, comparing the performance of the proposed ML-based approach to that of the perplexity-based approach, and by analyzing the explanations of model's decisions extracted using SHAP. \\

\begin{table*}[b]
\centering
\setlength{\tabcolsep}{5pt}
\caption{Perpexity- and ML-based classification results in terms of precision, recall, F1-score and accuracy for both experiments.}
\label{tab:QuantResults}
\begin{tabular}{cc|lllc|cccc}
\multicolumn{1}{l}{}    & \multicolumn{1}{l|}{} & \multicolumn{4}{l|}{Perplexity-based Classification}                                                    & \multicolumn{4}{c}{ML-based Classification}   \\ \cline{3-10}
                        &                       & \multicolumn{1}{c}{Prec.} & \multicolumn{1}{c}{Recall} & \multicolumn{1}{c}{F1} & Acc.                  & Prec. & Recall & F1   & Acc.                  \\ \hline
\multirow{2}{*}{Exp. 1} & Human                 & 0.83                      & 0.89                       & 0.86                   & \multirow{2}{*}{0.84} & 0.99  & 0.98   & 0.98 & \multirow{2}{*}{0.98} \\
                        & $ChatGPT_{query}$               & 0.85                      & 0.78                       & 0.81                   &                       & 0.97  & 0.99   & 0.98 &                       \\ \hline
\multirow{2}{*}{Exp. 2} & Human                 & 0.61                      & 0.72                       & 0.66                   & \multirow{2}{*}{0.69} & 0.71  & 0.80   & 0.75 & \multirow{2}{*}{0.79} \\
                        & $ChatGPT_{rephrase}$               & 0.78                      & 0.68                       & 0.72                   &                       & 0.85  & 0.77   & 0.81 &                       \\ \hline
\end{tabular}
\end{table*}

\textbf{\emph{RQ1) To what extent can ML-based approach and perplexity-based approach detect AI-generated text of relatively short length? And if ChatGPT-generated text is a rephrase of the original, human-generated, text?}}

Table \ref{tab:QuantResults} reports the experimental results obtained for the proposed ML-based classifier and the perplexity-based classifier for each of the two experiments conducted. Results show that classification accuracy of both approaches in Experiment 1 is higher (0.84 for perplexity-based approach and 0.98 for ML-based approach) than that in Experiment 2 (0.69 and 0.79, respectively). This is expected considering that rephrasing makes generated versions more similar to the original reviews, giving ChatGPT the possibility of maintaining human style writing, and thus making it easier to trick the classifier. Comparing the performance of the approaches in each of the experiments, we see that the ML-based approach significantly outperforms (achieves an accuracy of 0.98 in Experiment 1 and 0.79 in Experiment 2) the Perplexity-based approach (achieves an accuracy of 0.84 in Experiment 1 and 0.69 in Experiment 2). For Experiment 1, specifically, this shows that the perplexity-based approach, despite underperforming with respect to the ML-based one, could be considered as an acceptable approach to identify the source of a review. However, when AI-generated text is a rephrase of human-generated text (Experiment 2), perplexity alone is not sufficient for determining the review source and hence, cannot be considered a valid metric to use. This justifies the use of ML models, and in particular, Transformer-based models to classify review samples. 

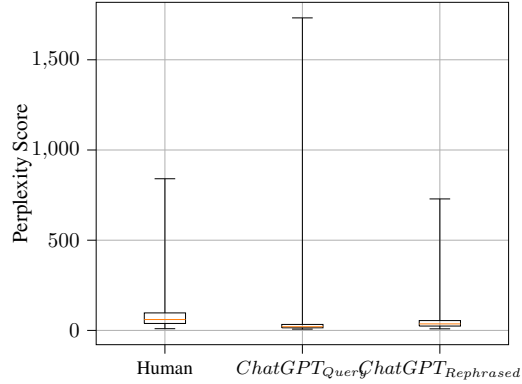
\begin{figure}{}
  \begin{center}
\begin{tikzpicture}[scale=0.8]

\definecolor{darkgray176}{RGB}{176,176,176}
\definecolor{darkorange25512714}{RGB}{255,127,14}

\begin{axis}[
tick align=outside,
tick pos=left,
x grid style={darkgray176},
xmajorgrids,
xmin=0.5, xmax=3.5,
xtick style={color=black},
xtick={1,2,3},
xticklabels={\small{Human},\small{$ChatGPT_{Query}$},\small{{$ChatGPT_{Rephrased}$}}},
y grid style={darkgray176},
ylabel={Perplexity Score},
ymajorgrids,
ymin=-79.1202965974808, ymax=1818.50969278812,
ytick style={color=black}
]
\addplot [black]
table {%
0.85 38.5211868286133
1.15 38.5211868286133
1.15 97.006742477417
0.85 97.006742477417
0.85 38.5211868286133
};
\addplot [black]
table {%
1 38.5211868286133
1 9.550217628479
};
\addplot [black]
table {%
1 97.006742477417
1 840.639831542969
};
\addplot [black]
table {%
0.925 9.550217628479
1.075 9.550217628479
};
\addplot [black]
table {%
0.925 840.639831542969
1.075 840.639831542969
};
\addplot [black]
table {%
1.85 14.3656444549561
2.15 14.3656444549561
2.15 33.1128826141357
1.85 33.1128826141357
1.85 14.3656444549561
};
\addplot [black]
table {%
2 14.3656444549561
2 7.13561201095581
};
\addplot [black]
table {%
2 33.1128826141357
2 1732.25378417969
};
\addplot [black]
table {%
1.925 7.13561201095581
2.075 7.13561201095581
};
\addplot [black]
table {%
1.925 1732.25378417969
2.075 1732.25378417969
};
\addplot [black]
table {%
2.85 24.1344728469849
3.15 24.1344728469849
3.15 54.4942855834961
2.85 54.4942855834961
2.85 24.1344728469849
};
\addplot [black]
table {%
3 24.1344728469849
3 8.78443336486816
};
\addplot [black]
table {%
3 54.4942855834961
3 728.806823730469
};
\addplot [black]
table {%
2.925 8.78443336486816
3.075 8.78443336486816
};
\addplot [black]
table {%
2.925 728.806823730469
3.075 728.806823730469
};
\addplot [darkorange25512714]
table {%
0.85 60.3485412597656
1.15 60.3485412597656
};
\addplot [darkorange25512714]
table {%
1.85 20.4698886871338
2.15 20.4698886871338
};
\addplot [darkorange25512714]
table {%
2.85 36.1514587402344
3.15 36.1514587402344
};
\end{axis}

\end{tikzpicture}
  \end{center}
 \caption{Box Plots of the perplexity values of the text samples of the three datasets considered in our analysis.}
    \label{fig:boxplot_figure}
\end{figure}

Figure \ref{fig:boxplot_figure} shows the box plots relative to the perplexity scores of the texts belonging to the two classes, i.e., generated by humans or by ChatGPT, used in each of the two experiments. 
Concerning Experiment 1, perplexity scores of texts written by humans are generally higher than those generated by ChatGPT (i.e., on average, the perplexity score of the human-generated texts is $84$, against $34$ of the ChatGPT-generated texts). We observe, however, that the maximum value of the human-generated texts is notably lower than the maximum value of the ChatGPT-generated texts (i.e., $840$ vs $1732$). Perplexity scores of texts of Experiment 2 show that the average perplexity value of the ChatGPT-generated text is lower than that of the human-generated texts (i.e., $47$ vs $84$). In this experiment, however, we note that the maximum value of perplexity for ChatGPT-generated text is much lower than that of Experiment 1, and is as well lower than that of human-generated text (i.e., $728$ vs $840$). This shows that using ChatGPT to rephrase human-generated text can yield lower perplexity scores of the generated text, and hence, makes it harder to detect that the text is generated by an AI (in this case, by ChatGPT-2). \\

\textbf{\emph{RQ2) What are the characteristics of text that contribute to detecting an AI-generated text? Can we extract insights by examining explanations of ML model decisions?}}

To address RQ2, we focus on explaining the outcomes of the ML model and on analyzing the extracted explanations with the aim of understanding what the ML model basis its decisions on when claiming that a text is generated by ChatGPT. Put simple, we aim to identify if there are specific words or patterns of words specific to text generated by ChatGPT that the ML model perceives and, based on which, identifies a text as ChatGPT-generated. 

We extract local explanations using SHAP. For each of the two experiments, we consider a random train-test split of the dataset and extract local explanations of all test points (i.e., of all text samples in the test set). We explain test points towards the target class (i.e., towards ChatGPT-generated text). Positive SHAP values indicate that the feature had a positive impact on the prediction (in this case, positive impact towards classifying text as ChatGPT-generated text), while negative values indicate that the feature had a negative impact. The magnitude of the SHAP value represents the degree of impact the feature had on the prediction. The local explanation plot shows the feature names on the y-axis (listed from top to bottom based on their impact on decision), and the x-axis represents the SHAP values for each feature. 

\begin{figure}[h]{
\subfigure[]
{
\includegraphics[width=0.33\textwidth, height = 4.8cm]{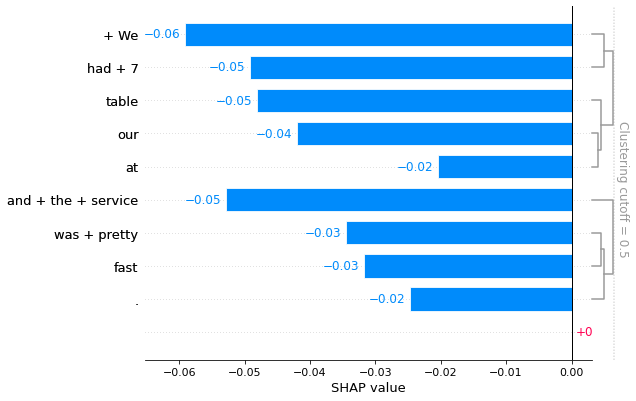}
}
\subfigure[]
{
\includegraphics[width=0.33\textwidth, height = 4.8cm]{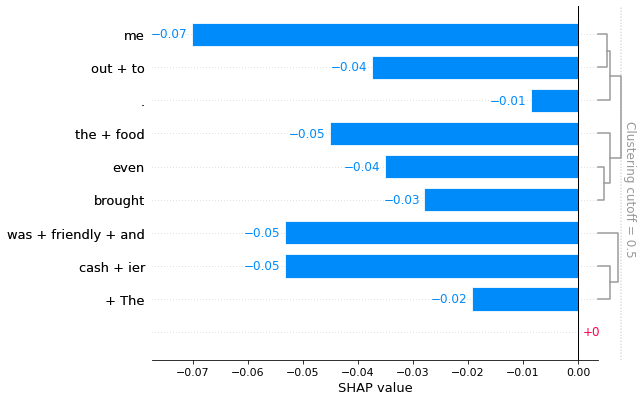}
}
\subfigure[]
{
\includegraphics[width=0.33\textwidth, height = 4.8cm]{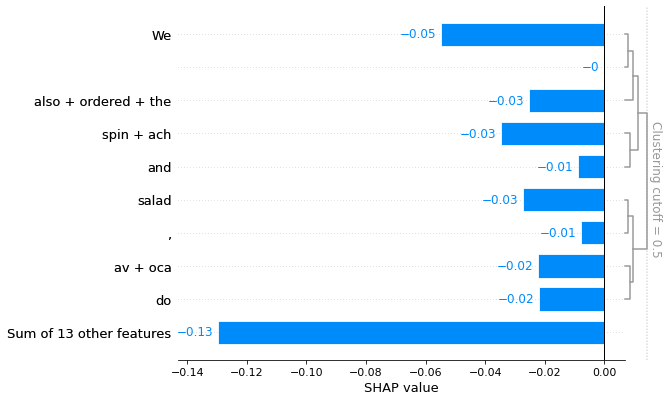}
}
\caption{SHAP local explanation plots for three decisions corresponding to three different text samples (data points): (a) \emph{``We had 7 at our table and the service was pretty fast.''}, (b) \emph{``The cashier was friendly and even brought the food out to me.''}, (c) \emph{``We also ordered the spinach and avocado salad, the ingredients were sad and the dressing literally had zero taste.''}}
\label{fig:exp1_1a}
}
\end{figure}

\begin{figure}[h]{
\subfigure[]
{
\includegraphics[width=0.33\textwidth, height = 4.8cm]{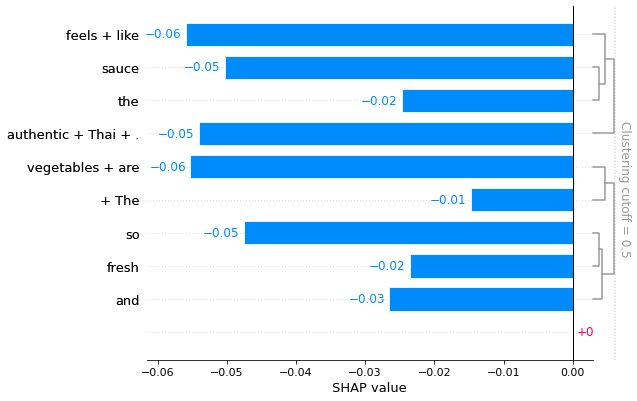}
}
\subfigure[]
{
\includegraphics[width=0.33\textwidth, height = 4.8cm]{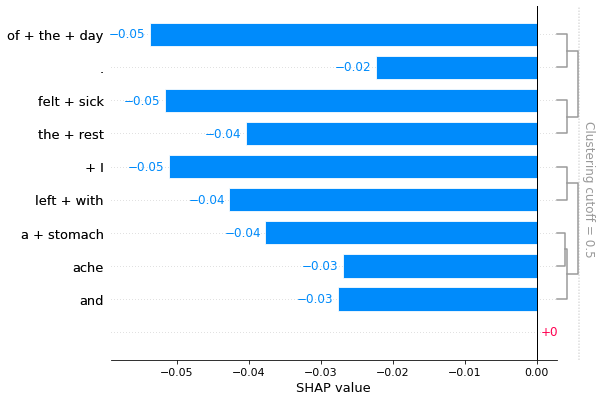}
}
\subfigure[]
{
\includegraphics[width=0.33\textwidth, height = 4.8cm]{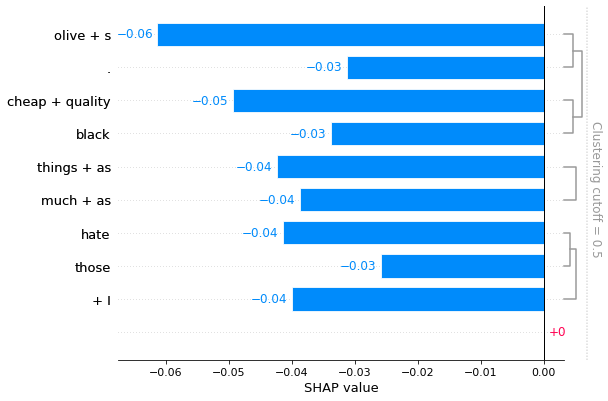}
}
\caption{SHAP local explanation plots for three decisions corresponding to three different text samples (data points): (a) \emph{``The vegetables are so fresh and the sauce feels like authentic Thai.''}, (b) \emph{``I left with a stomach ache and felt sick for the rest of the day.''}, (c) \emph{``I hate those things as much as cheap quality black olives.''}}
\label{fig:exp1_1b}
}
\end{figure}

\textbf{Experiment 1} In Experiment 1, we have selected a subset of explanations for further examination. Figure \ref{fig:exp1_1a} illustrates three explanation plots for decisions in which the model's outcome is classified as human-generated text. It is observed that personal pronouns, such as \emph{``We''}, \emph{``me''}, and \emph{``our''}, have a relatively high impact on the decision, in this case, negatively contributing to the classification of ChatGPT-generated text. In Figure \ref{fig:exp1_1b}, we present explanations for three additional samples which the model also classifies as human-generated text, but the reasons for doing so differ from those in Figure \ref{fig:exp1_1a}. Specifically, it is noted that words expressing feelings, such as \emph{``feels like''}, \emph{``felt sick''} and \emph{``hate''}, have a negative impact on the classification of ChatGPT-generated text, thus positively contributing to the detection of human-generated text. These findings indicate that the machine learning model (in this case, the one trained on the data set of Experiment 1) is able to effectively correlate input features related to feelings with human-generated text.

\begin{figure}[h]{
\subfigure[]
{
\includegraphics[width=0.33\textwidth, height = 4.8cm]{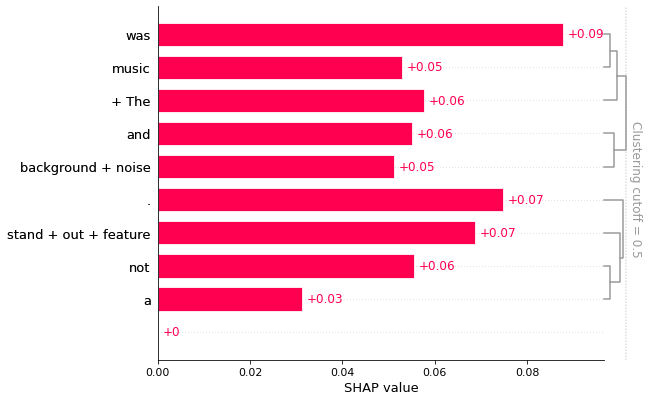}
}
\subfigure[]
{
\includegraphics[width=0.33\textwidth, height = 4.8cm]{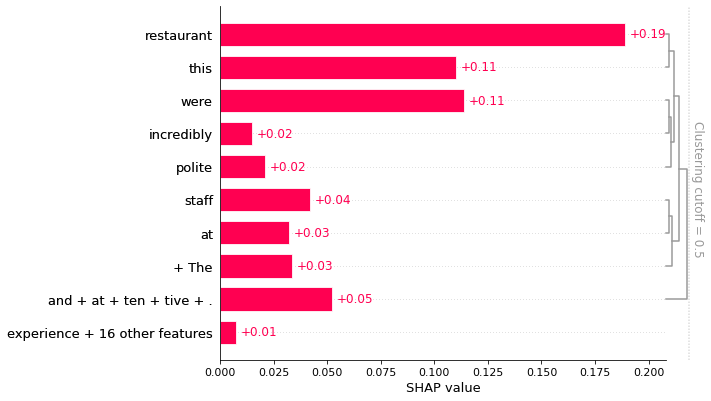}
}
\subfigure[]
{
\includegraphics[width=0.33\textwidth, height = 4.8cm]{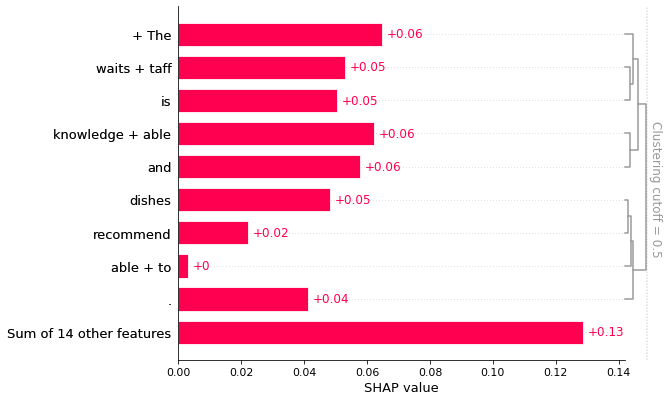}
}
\caption{SHAP local explanation plots for three decisions corresponding to three different text samples (data points): (a) \emph{``The music was background noise and not a standout feature.''}, (b) \emph{``The staff at this restaurant were incredibly polite and attentive. They made sure we had everything we needed and made our dining experience a pleasure.''}, (c) \emph{``The waitstaff was knowledgeable and able to recommend dishes.''}}
\label{fig:exp1_2}
}
\end{figure}

Figure \ref{fig:exp1_2} depicts three local explanations plots which showcase the model's outcomes for decisions related to ChatGPT-generated text. The SHAP values of the features (words) in these explanations are positive, indicating that they positively contributed to the classification of the text as ChatGPT-generated. Upon analysis of these explanations, it is evident that the model has learned to correlate certain words, such as \emph{``stand out feature''}, \emph{``incredibly polite''}, \emph{``waitstaff''} and \emph{``knowledgeable''}, with ChatGPT-generated text. These words are not commonly used by humans in restaurant reviews, providing further evidence of the model's ability to accurately distinguish between human-generated and ChatGPT-generated text.

\begin{figure}[h]{
\centering
\subfigure[]
{
\includegraphics[width=0.33\textwidth, height = 4.8cm]{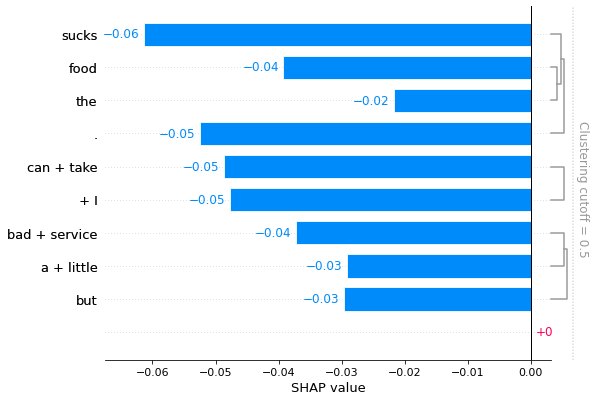}
}
\subfigure[]
{
\includegraphics[width=0.33\textwidth, height = 4.8cm]{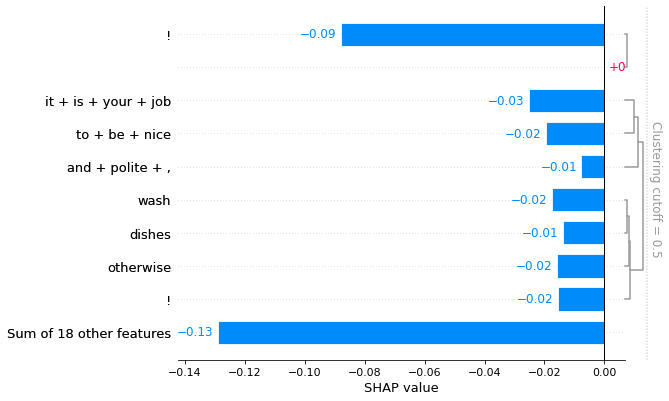}
}
\caption{SHAP local explanation plots for two decisions corresponding to two different text samples (data points): (a) \emph{``I can take a little bad service but the food sucks.''}, (b) \emph{``Bad day or not, I have a very low tolerance for rude customer service people, it is your job to be nice and polite, wash dishes otherwise!''}}
\label{fig:exp1_3}
}
\end{figure}

As depicted in Figure \ref{fig:exp1_3}, the three local explanations plots presented demonstrate the model's predictions for human-generated text, with a notable emphasis on impolite expressions. The first plot illustrates that the word \emph{``sucks''} has the greatest influence on the model's decision, which is in alignment with human-generated text. Similarly, in the second plot, the word \emph{``shoe leather''} has the largest impact on the model's decision. These findings indicate that the machine learning model is able to effectively correlate aggressive and impolite language with human writing, as opposed to ChatGPT-generated text.

Observations from Experiment 1:
\begin{itemize}
\setlength\itemsep{0em}
    \item \textbf{ChatGPT tends to describe experiences rather than expressing feelings}. In several cases, expression of feelings was attributed to human-generated text class.
    \item \textbf{ChatGPT, unlike humans, refrains from using personal pronouns}, which are, in most cases, attributed to human-generated text class. 
    \item \textbf{ChatGPT tends to use uncommon (unusual) words}. In most cases, such words attributed to detecting the ChatGPT-generated text class.
    \item \textbf{Aggressive language and rude vocabs are never used by ChatGPT}. The ML model attributes the use of such vocabs to human-generated text class. 
\end{itemize}

\textbf{Experiment 2} We now concentrate on the set of explanations of Experiment 2. We first note that similar observations as those found in Experiment 1 can be extracted from explanations of Experiment 2, however with less occurrence (and hence, less evident). This is to be expected, as in Experiment 2, ChatGPT is asked to rephrase human-generated text and thus has less flexibility in terms of word choice and writing style. In fact, the results of Experiment 2 demonstrate a decrease in model accuracy, with a score of 0.79, compared to Experiment 1, which recorded an accuracy of 0.98. This discrepancy highlights the difficulty in accurately distinguishing between ChatGPT- and human-generated text. In order to gain further insight into this issue, we have undertaken an analysis of the model's misclassifications, with a particular focus on identifying the factors that led to not detecting ChatGPT-generated text.

\begin{figure}[h]{
\centering
\subfigure[]
{
\includegraphics[width=0.31\textwidth, height = 4.8cm]{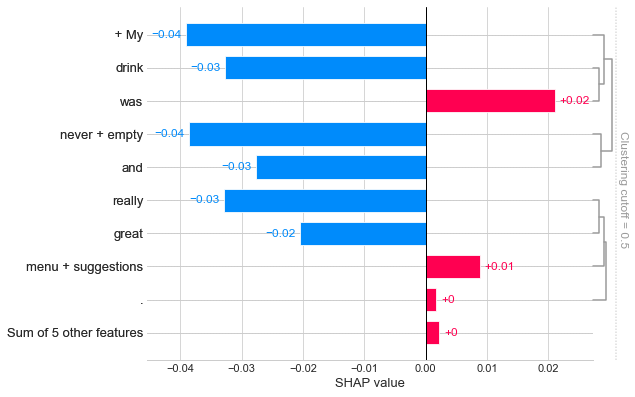}
}
\subfigure[]
{
\includegraphics[width=0.32\textwidth, height = 4.8cm]{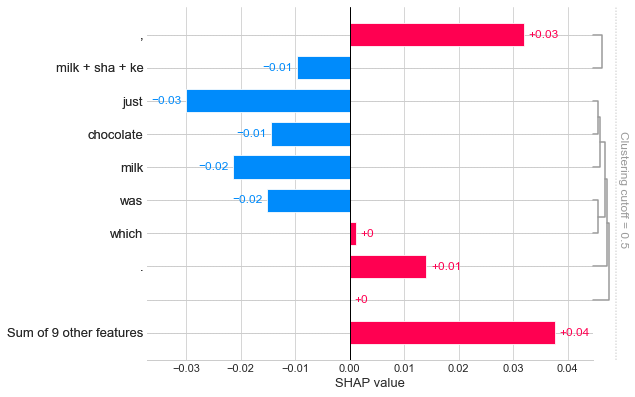}
}
\caption{SHAP local explanation plots for two misclassified text samples (data points) by the model: (a) \emph{``My drink was never empty and he made some really great menu suggestions.''} , (b) \emph{``It took over 30 min to get their milkshake, which was nothing more than chocolate milk.''}}
\label{fig:exp2_1}
}
\end{figure}

Figure \ref{fig:exp2_1} presents the local explanation plots of misclassifications in which the ground truth is text generated by ChatGPT, while the predicted text is human-generated. In Figure \ref{fig:exp2_1}(a), it is observed that most words contribute to the classification as human-generated text. In particular, the personal pronoun \emph{``My''} exhibits the highest value in the SHAP contribution analysis. Similarly, Figure \ref{fig:exp2_1}(b) illustrates that the words \emph{``milkshake''}, \emph{``chocolate''}, and \emph{``milk''} also contribute to the classification as human-generated text. This may correspond to a scenario in which the text describes specific details about a dining experience, which is a common characteristic of the human-generated text. These examples demonstrate the difficulties in identifying text generated by ChatGPT when it is based on rephrasing original human-generated text.

\section{Discussion}
\label{sec:Discussion}

The obtained classification results clearly show that the disambiguation between ChatGPT-generated and human-generated reviews is more challenging for the ML model when the reviews have been rephrased from existing human-generated texts and not generated by custom queries. Yet, the ML model is capable to achieve an acceptable performance with an accuracy around 79\%. We further note that the approach based on rephrasing existing reviews, besides being more effective in avoiding detection of ChatGPT-generated text for relatively short-length text (such as online reviews), it is also more scalable, as it eliminates the need of crafting custom queries to obtain the reviews. This fact opens the doors for possible misuse of ChatGPT, which might be used to easily obtain plenty of negative reviews that can mislead consumers and damage the reputation of businesses or, on the other hand, to improve the reputation of low-quality businesses. 

Regarding our analysis of local explanations of model's decisions, we need to emphasize that the explanations are data dependent, and we would need a more extensive analysis before we could claim being able to discriminate between ChatGPT and human writing styles in every detail. Nevertheless, the obtained Transformer-based classification model performance (which turned out to be much better than what we expected!) and obtained local explanations, demonstrate that even from this preliminary study we could notice some kind of patterns in the writing styles. First, as mentioned, ChatGPT language is impersonal (no personal pronouns, no expressions of feelings, but instead, describing experience and using mostly the mixture of third tense and passive speech), unless, of course, asked to. 
Second, ChatGPT quite repeats itself. A majority of reviews in the $ChatGPT_{query}$ dataset, starts with \emph{``the restaurant''}, \emph{``this restaurant''} or contains the word \emph{``restaurant''}. If by a chance it brings up a specific restaurant name, it will keep repeating it for the next reviews as well. In addition, if asked to produce more reviews, oftentimes the following ones will be obtained by rephrasing or just shuffling sentence parts around. Third, ChatGPT reviews are more general and content-wise more ``common'', e.g. it can speak about quality of service, food, atmosphere in general, but humans go into much more specific details of e.g. describe their meals (see Figure \ref{fig:exp1_1a}(c)), particularities about service (see Figure \ref{fig:exp1_1a}(b)), sitting arrangements (see Figure \ref{fig:exp1_1a}(a)) etc. Forth, ChatGPT vocabulary is much more formal (e.g. \emph{``noteworthy''}) and misses colloquial terms and abbreviations (e.g. it never uses ``\&'' instead of \emph{``and''}). On the other hand, it could contain atypical words or language constructs (e.g. \emph{``inattentive''}), or simply be overreacting and exaggerating, especially in positive cases (e.g. \emph{``absolutely delicious''}, \emph{``incredibly polite''}). Fifth, ChatGPT as rightfully their authors claim, does not use inappropriate language (neither offensive nor hate speech). However, at least from what we were able to see, it also does not use metaphors, irony or sarcasm. 



\section{Conclusion}\label{sec:Conc}

In this paper we focus on building ML (Transformer-based) model to discriminate between human-written and seemingly human (ChatGPT-generated) text, focusing on a more challenging case, short texts. We show that the usage of ML model is necessary as traditional methods (such as, perplexity) do not give good results. As expected, ML model discriminates better when the text is generated based on customer queries and not by rephrasing original human texts. Looking into SHAP explanations of the predictions gives some insights about ChatGPT writing style. It is extremely polite, aiming to please different types of requests from various domains fairly well mimicking humans, but that still does not have the profoundness of human language (e.g. irony, metaphors,...). For the future work, we will consider alternative ML models, different domains and query types.

\bibliographystyle{splncs03}
\bibliography{references}
\end{document}